\pdfoutput=1

\documentclass[11pt]{article}

\usepackage[final]{coling}

\usepackage{times}
\usepackage{latexsym}

\usepackage[T1]{fontenc}

\usepackage[utf8]{inputenc}

\usepackage{microtype}

\usepackage{inconsolata}

\usepackage{graphicx}

\usepackage{soul}
\usepackage{adjustbox}
\usepackage{hyperref}
\usepackage{multirow}
\usepackage{booktabs}
\usepackage{dirtytalk}
\usepackage{array}
\usepackage{subcaption}
\usepackage{caption}
\usepackage{amsmath}
\usepackage{float}
\usepackage{longtable}

\title{Unmasking the Imposters: How Censorship and Domain Adaptation Affect the Detection of Machine-Generated Tweets}

  \author{Bryan E. Tuck \\
  University of Houston \\
  Houston, TX, USA \\
  \texttt{betuck@uh.edu} \\\And
  Rakesh M. Verma \\
  University of Houston \\
  Houston, TX, USA \\
  \texttt{rmverma2@central.uh.edu} \\}

\begin{document}

\maketitle

\begin{abstract}

The rapid development of large language models (LLMs) has significantly improved the generation of fluent and convincing text, raising concerns about their potential misuse on social media platforms. We present a comprehensive methodology for creating nine Twitter datasets to examine the generative capabilities of four prominent LLMs: Llama 3, Mistral, Qwen2, and GPT4o. These datasets encompass four censored and five uncensored model configurations, including 7B and 8B parameter base-instruction models of the three open-source LLMs. Additionally, we perform a data quality analysis to assess the characteristics of textual outputs from human, \say{censored,} and \say{uncensored models,} employing semantic meaning, lexical richness, structural patterns, content characteristics, and detector performance metrics to identify differences and similarities. Our evaluation demonstrates that uncensored models significantly undermine the effectiveness of automated detection methods. This study addresses a critical gap by exploring smaller open-source models and the ramifications of \say{uncensoring,} providing valuable insights into how domain adaptation and content moderation strategies influence both the detectability and structural characteristics of machine-generated text.

\end{abstract}

\section{Introduction} 
\label{sec:introduction}
Maintaining the integrity of digital communication platforms like Twitter has become increasingly vital due to exponential advancements in large language models (LLMs) in recent years. While the misuse of technology by \say{bad actors} is not new, the scale at which they can now disseminate misinformation, hate speech, and convincingly imitate others has grown alarmingly \cite{wu2024survey}. This development poses significant challenges distinguishing between human and machine-generated content, especially on social media, where generative AI can have far-reaching consequences \cite{Dhaini2023DetectingCA}.

The ongoing advancements in LLMs are fascinating due to their transformative potential, yet they pose significant risks. These models can generate highly convincing misinformation and fake news on an unprecedented scale, undermining the trust and reliability of digital communication platforms \cite{Jawahar2020AutomaticDO, Crothers2022MachineGeneratedTA}. Consequently, it is critical to develop robust automated detection systems to identify and mitigate the spread of false information. By addressing these challenges, we can safeguard digital spaces, ensure informed public discourse, and mitigate the harmful impacts of generative AI.

However, the growing sophistication of LLMs has made detecting machine-generated text increasingly difficult \cite{DaSilvaGameiro2024}. Naive approaches often fail because LLMs can incorporate recent information and adapt to specific writing styles \cite{Liu2024CustomizingLL, Buz2024InvestigatingWC}. Additionally, the availability of small, efficient, open-source LLMs allows \say{bad actors} to fine-tune models on specific domains, producing highly fluent and convincingly human-like text tailored to particular contexts.

Previous methods for detecting machine-generated text often focus on general-purpose datasets and models, which may not capture the specific traits of social media text like informal language, short messages, and emojis. These approaches typically rely on well-known models, such as GPT variants \cite{openai2023chatgpt}, or \say{nano} LLMs (fewer than 1.5 billion parameters), lacking domain adaptation. They also overlook the increasing use of open-source community models \cite{Kumarage2023StylometricDO, Alamleh2023DistinguishingHA, Abburi2023GenerativeAT, Nguyen2023StackingTO, Lai2024AdaptiveEO, Wang2023BotOH}. Our work expands on this by evaluating a wider range of LLMs, including both censored and uncensored models, as well as GPT4o, on domain-specific social media data, offering a more realistic perspective on identifying machine-generated content from \say{bad actors.}

To address these challenges, we use Twitter datasets from TweetEval \cite{barbieri-etal-2020-tweeteval} to differentiate between human and machine-generated tweets. Our study focuses on four main LLMs: Llama 3 \cite{Touvron2023Llama2O}, Mistral \cite{Jiang2023Mistral7}, Qwen2 \cite{qwen2}, GPT4o \cite{openai2023chatgpt}, and five additional community-driven uncensored models, resulting in nine unique models with varying content moderation and fine-tuning. We analyze these models based on semantic meaning, lexical richness, structure, and content to identify differences between censored and uncensored generative models. Additionally, we evaluate the generated tweets using detection methods like BERTweet \cite{Nguyen2020BERTweetAP}, DeBERTa \cite{he2023debertav}, a soft-voting ensemble, and stylometric features \cite{Kumarage2023StylometricDO}. Though limited to Twitter, this study offers key insights into the strengths and limitations of censored and uncensored open-source LLMs in generating domain-specific content and detecting machine-generated text in real-world applications.

\subsection*{Summary of Contributions} 
\begin{enumerate} 
    \item \textbf{Novel Evaluation Framework}: We introduce a methodology for adapting public Twitter datasets to assess both \textit{censored} and \textit{uncensored} state-of-the-art LLMs, addressing a research gap focused primarily on GPT models (Section \ref{sec:methodology}).

    \item \textbf{Comprehensive Quality Analysis}: By examining semantics, lexical richness, structure, toxicity, and detector performance, we reveal how removing safety moderation enables models to produce more human-like yet potentially more harmful text (Section \ref{sec:sem_lex_struc}).

    \item \textbf{Benchmark and Detector Insights}: We provide nine benchmark subsets and show that standard detectors deteriorate against \textit{uncensored} models, offering a foundation for future improvements in detection and moderation approaches (Section \ref{sec:detector_perf}).
\end{enumerate}

\section{Related Work}
\label{sec:related_work}
\subsection*{Stylometric and Machine Learning Approaches} 
Stylometry and machine learning have been employed to automate the detection of machine-generated fake news and text. \citet{Schuster2019TheLO} demonstrated the effectiveness of stylometry in identifying text origin but highlighted its limitations in distinguishing legitimate and malicious uses of language models. \citet{Bakhtin2019RealOF} showed that energy-based models exhibit good generalization across different generator architectures but are sensitive to the training set. \citet{Kumarage2023StylometricDO} proposed a novel algorithm using stylometric signals to detect AI-generated tweets generated by GPT2 and EleutherAI-gpt-neo-1.3B \cite{gao2020pile}, showing that stylometric features can effectively augment state-of-the-art detectors in Twitter timelines or limited training data.

Recent studies have also tackled differentiating human-written and AI-generated text in academic contexts. \citet{Alamleh2023DistinguishingHA} demonstrated the high accuracy of machine learning models, especially random forests and SVMs, in this task. \citet{Abburi2023GenerativeAT} introduced an ensemble neural model that leverages probabilities from pre-trained language models as features, yielding strong performance in binary and multi-class classification across English and Spanish. \citet{Nguyen2023StackingTO} showcased the power of ensembling lightweight transformers, achieving 95.55\% accuracy on a shared task test set. However, \cite{Lai2024AdaptiveEO} observed that while single transformer-based models excel on in-distribution data, they struggle with out-of-distribution samples.

\subsection*{Zero-shot and Few-shot Detection Methods} 
Zero-shot and few-shot detection methods have shown promise in identifying machine-generated text. DetectGPT \cite{Mitchell2023DetectGPTZM} leverages the curvature of a language model's log probability function to outperform existing baselines without additional training. FLAIR \cite{Wang2023BotOH} uses carefully designed questions to elicit distinct responses from bots and humans, proving effective in differentiating between the two in an online setting.

\citet{Mireshghallah2024SmallerLM} showed that smaller language models are more effective at detecting machine-generated text, regardless of the generator's architecture or training data. \citet{Mitrovic2023ChatGPTOH} explored how machine learning models distinguish between human-generated and ChatGPT-generated text in short online reviews, identifying patterns like polite language, lack of specific details, and impersonal tone.

\subsection*{Adversarial Attacks and Defenses} 
AI-generated text detectors are vulnerable to adversarial attacks, particularly those involving paraphrasing. \citet{Krishna2023ParaphrasingED} introduced DIPPER, an 11b paraphrase generation model that can evade several detectors, including watermarking, GPTZero, DetectGPT, and OpenAI's now defunct text classifier \cite{Tian2023, Mitchell2023DetectGPTZM, Kirchner2023NewAI}. \citet{Kumarage2023HowRA} challenged the reliability of current state-of-the-art detectors by introducing EScaPe, a framework that learns evasive soft prompts that guide pre-trained language models to generate text that deceive detectors.

To improve AI-generated text detection, \cite{Hu2023RADARRA} proposed RADAR, which jointly trains a detector and a paraphraser via adversarial learning, significantly outperforming existing methods. Finally, \cite{Sadasivan2023CanAT} revealed reliability issues with watermarking, neural network-based, zero-shot, and retrieval-based detectors by developing a recursive paraphrasing attack that compromises watermarking and retrieval-based detectors with minimal text quality degradation, exposing their vulnerability to spoofing attacks.

\subsection*{Datasets and Benchmarks} 
Large-scale datasets are important for developing effective machine-generated content detection algorithms. \citet{Fagni2020TweepFakeAD} introduced TweepFake, the first dataset of deepfake tweets based on GPT 2, recurrent neural networks, and Markov Chains, benchmarking traditional machine learning, character convolutional networks, and Bert-based models. \citet{Yu2023CHEATAL} introduced CHEAT, a dataset containing 35,304 ChatGPT-generated abstracts, analyzing the distribution differences between human-written and ChatGPT-written abstracts. \citet{Li2023DeepfakeTD} constructed a testbed for deepfake text detection, collecting human-written texts from diverse domains and generating corresponding deepfake texts using GPT 3.5, T5, and Llama.

Recent efforts have focused on advancing multilingual machine-generated text detection. \citet{macko-etal-2023-multitude} introduced MULTITuDE, a benchmark of over 74,000 texts across 11 languages, highlighting the challenges detectors face when generalizing to unseen languages and language families. Building on this, M4 \cite{wang-etal-2024-m4} expanded detection to multi-domain, multi-generator, and multilingual settings, demonstrating that detectors often struggle with unfamiliar domains and LLMs. In a continuation, M4GT-Bench \cite{wang-etal-2024-m4gt} broadened the scope by covering nine languages, six domains, and nine state-of-the-art LLMs while introducing novel tasks like mixed human-machine text detection.

\section{Methodology}
\label{sec:methodology}

\subsection{Datasets}
\label{sec:datasets}
We use the TweetEval unified benchmark \cite{barbieri-etal-2020-tweeteval} for our human-labeled tweets. Specifically, we extract the emotion, irony, sentiment, hate speech, and offensive language datasets for fine-tuning our LLMs. We use the emotion recognition subset to generate our synthetic tweets. Detailed distributions are in Table \ref{tab:tweeteval_datasets}. 

\begin{table}[H]
\centering
\begin{adjustbox}{max width=\columnwidth}
\begin{tabular}{l|c|c|c|c|c}
\hline
\textbf{Task} & \textbf{Labels} & \textbf{Train} & \textbf{Val} & \textbf{Test} & \textbf{Total} \\
\hline
Emotion recognition & 4 & 3,257 & 374 & 1,421 & 5,052 \\
Hate speech detection & 2 & 9,000 & 1,000 & 2,970 & 12,970 \\
Irony detection & 2 & 2,862 & 955 & 784 & 4,601 \\
Offensive language identification & 2 & 11,916 & 1,324 & 860 & 14,100 \\
Sentiment analysis & 3 & 45,615 & 2,000 & 12,284 & 59,899 \\
\hline
Total & 13 & 72,650 & 5,653 & 18,319 & 96,622 \\
\hline
\end{tabular}
\end{adjustbox}
\caption{TweetEval datasets we use for fine-tuning our large language models for domain adaptation to Twitter.}
\label{tab:tweeteval_datasets}
\end{table}

\subsection{Large Language Models}
\begin{table}[H]
\centering
\begin{adjustbox}{max width=\columnwidth}
\begin{tabular}{@{}lcl>{\centering\arraybackslash}p{3cm}@{}}
\toprule
\textbf{Model} & \textbf{Parameters} & \textbf{Variant} & \textbf{Abbreviation} \\ \midrule
Llama 3 & 8B & Censored: Meta-Llama-3-8B-Instruct & LL3 \\ 
        &    & Uncensored: Dolphin-2.9-Llama 3-8B & LL3-Dolphin \\
        &    & Uncensored: Hermes 2 Pro-Llama-3-8B & LL3-Hermes\\ \midrule
Mistral & 7B & Censored: Mistral-7B-Instruct-v0.2 & Mistral \\
        &    & Uncensored: Dolphin-2.8-Mistral-7B-v02 & Mistral-Dolphin \\
        &    & Uncensored: OpenHermes-2.5-Mistral-7B & Mistral-Hermes \\ \midrule
Qwen2   & 7B & Censored: Qwen2-7B-Instruct & Qwen2 \\
        &    & Uncensored: Dolphin-2.9.2-Qwen2-7B & Qwen2-Dolphin \\ \midrule
GPT-4o  & Closed-source & Censored & GPT-4o \\ \bottomrule
\end{tabular}
\end{adjustbox}
\caption{Overview of the large language models (LLMs) and their variants, including parameter sizes and the corresponding abbreviations used in this study. We look into the diversity of four LLM architectures (Llama 3, Mistral, Qwen2, and GPT-4o) and distinguish between censored and uncensored versions for subsequent analyses.}
\label{tab:llm_compare}
\end{table}

\begin{figure}[h]
    \centering
    \includegraphics[width=0.45\textwidth]{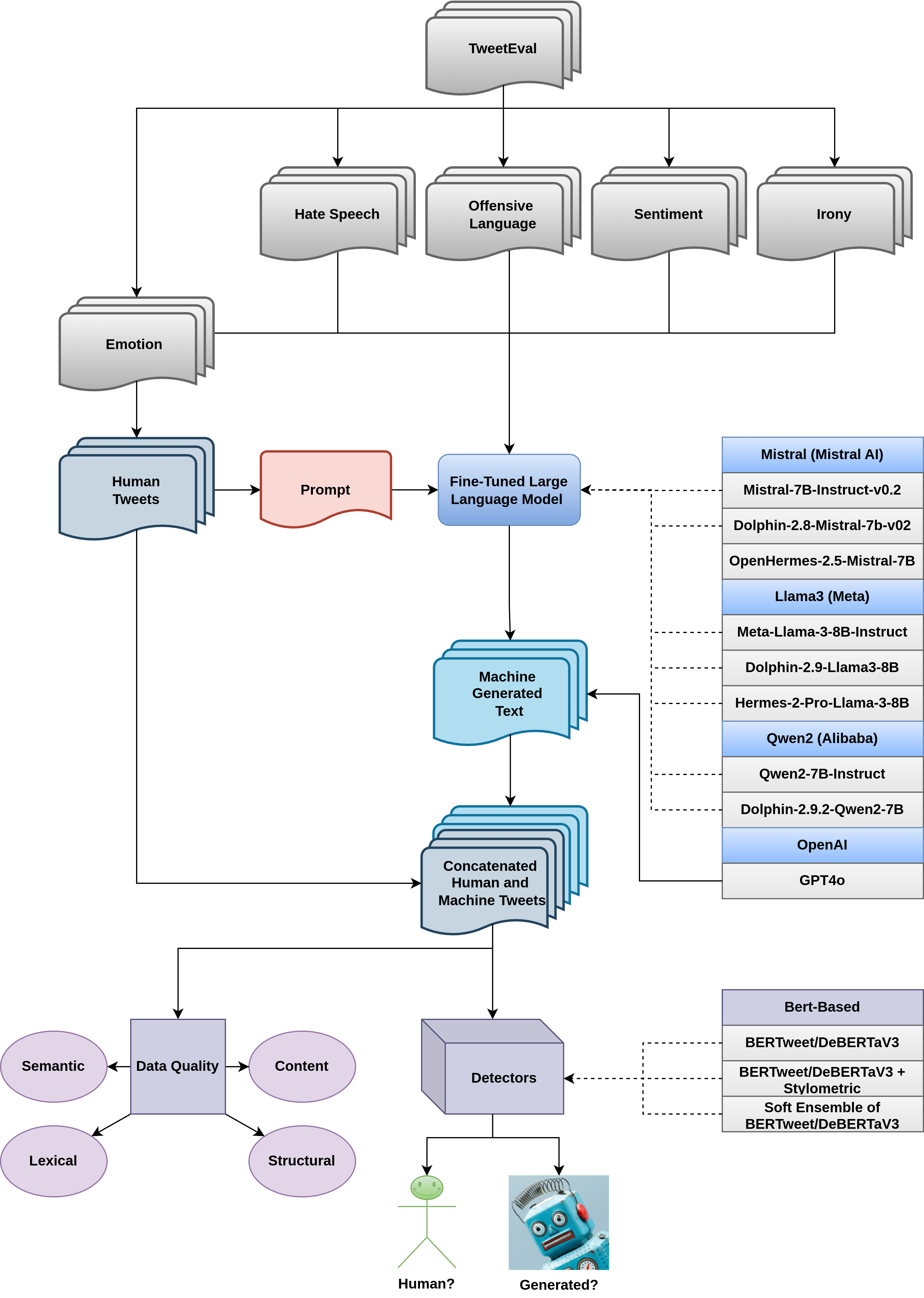}
    \caption{Overview of the experimental workflow. We begin with the TweetEval benchmark, selecting tweets from multiple tasks (emotion, hate speech, offensive language, sentiment, and irony) to fine-tune our LLMs. The emotion dataset tweets serve as prompts for the fine-tuned large language models (LLMs), which generate synthetic, domain-specific text. The resulting human and machine-generated tweets are combined, assessed for data quality across semantic, lexical, structural, and content dimensions, and finally evaluated by our six detection methods (including BERT-based models, stylometric features, and ensembles) to determine whether each sample is human or machine-produced.}
    \label{fig:workflow}
\end{figure}

We employ four primary large language models (LLMs) families: Llama 3 (8B) \cite{Touvron2023Llama2O}, Mistral (7B) \cite{Jiang2023Mistral7}, Qwen2 (7B) \cite{qwen2}, and the closed-source GPT-4o \cite{openai2023chatgpt}. These models were selected for their robust performance across diverse natural language processing tasks, balanced by their relatively small parameter sizes and accessibility without requiring extensive computational resources. Figure \ref{fig:workflow} outlines our experimental workflow, while Table \ref{tab:llm_compare} presents a comparative analysis of the model variants, including abbreviations we will use in the following sections.

Meta’s Llama 3 (8B) offers an optimal balance between computational efficiency and fine-tuning capability; we use three variants: the censored Meta-Llama-3-8B-Instruct, uncensored Hermes 2 Pro-Llama-3-8B \cite{Hermes-2-Pro-Llama-3-8B} fine-tuned on the OpenHermes 2.5 dataset \cite{OpenHermes-2.5}, and the uncensored Dolphin-2.9-Llama-3-8B aimed at reducing alignment biases \cite{ding2023enhancing}. 

Mistral (7B) is recognized for surpassing larger models like Llama 2 (13B) on benchmark tests. We include the censored instruction-tuned Mistral-7B-Instruct-v0.2, the uncensored Dolphin-2.8-Mistral-7B-v02 variant, and the code-optimized OpenHermes-2.5-Mistral-7B model.

Qwen2 (7B) advances multilingual capabilities by supporting 27 additional languages beyond English and Chinese. It is evaluated in its censored base-instruct and the uncensored Dolphin-2.9.2-Qwen2-7B variant to assess the impact of censorship across different languages and tasks. Lastly, GPT-4o serves as a benchmark to compare open-source models against proprietary technology, facilitating an analysis of performance, safety, and bias control.

This diverse selection enables examining content moderation effects and the trade-offs between model performance and responsible AI development. Links to each open-source model weights and documentation are provided in appendix \ref{sec:model_links}, Table \ref{tab:model_links}.

\subsubsection{Fine-tuning on In-Domain Twitter Data}
We adapt our LLMs to social media text by fine-tuning them on the concatenated TweetEval datasets (Table \ref{tab:tweeteval_datasets})—only using tweets longer than ten characters, resulting in 96,225 tweets, split into 91,413 for training and 4,817 for validation.

To maintain the authenticity of the text, we apply minimal preprocessing, masking only user mentions and URLs. Our fine-tuning process, implemented with the Hugging Face Transformers library \cite{Wolf2019TransformersSN} and the PEFT framework, using 4-bit Quantized Low-Rank Adaptation (QLoRA) \cite{hu2022lora, Dettmers2023QLoRAEF} to reduce the number of trainable parameters while preserving model performance. By leveraging the diverse tweet types in the TweetEval datasets, we aim to improve the models' ability to generate more relevant, high-quality tweets. Our trainer and QLoRA configuration are detailed in Table \ref{tab:trainer_params} and Table \ref{tab:lora_params} in Appendix \ref{sec:qlora_finetune}.

\subsubsection{Synthetic Tweet Generation}
We generate synthetic tweets using the original emotion recognition dataset as a foundation. For each human tweet, we create a prompt that instructs the language model to produce a new tweet conveying the same emotion. The prompt incorporates the human tweet and specific generation guidelines, such as employing creative and diverse linguistic techniques, substituting words or phrases with semantically similar alternatives, and varying sentence structures. We utilize the default greedy decoding method, as preliminary experiments indicated it produces more nuanced and harder-to-detect text than beam or contrastive search strategies.

The generation process follows a structured dialogue between the user and the AI assistant, as illustrated in Table \ref{tab:prompt_interaction}. Initially, the user provides the human tweet and generation guidelines, explicitly specifying the desired emotion to ensure consistency with the original. The assistant acknowledges the request and prompts for the tweet content. This explicit specification enables the language model to focus on accurately reflecting the same emotional content while introducing linguistic variations and diversity.
\begin{table}[h]
\centering
\begin{adjustbox}{max width=0.47\textwidth}
\begin{tabular}{|l|p{6cm}|}
\hline
\textbf{Role} & \textbf{Interaction and Prompt} \\
\hline
User & "You are an AI assistant tasked with crafting tweets that convey the same emotion as the original tweet, which is \{original\_emotion\}." \\
\hline
Assistant & "Understood. Please share the original tweet. I'll generate a version that follows the provided guidelines." \\
\hline
User & "Original tweet: \{text\}\newline
Generate a complete tweet following these guidelines:\newline
- Expresses the same emotion as the original tweet (\{original\_emotion\}) using creative and diverse linguistic techniques.\newline
- Substitutes the original tweet's words/phrases with semantically similar alternatives and varies the sentence structure.\newline
- Only return the generated tweet, refraining from using 'here's a tweet that conveys'.\newline
Generated tweet:" \\
\hline
\end{tabular}
\end{adjustbox}
\caption{Sample prompt and response template illustrating our interactive generation process. The user first establishes the desired emotion and requests a reformulated tweet, while the assistant acknowledges and requests the original content. The user then provides the initial tweet and detailed transformation guidelines. This structured exchange guides the LLM in producing a new, emotionally aligned tweet that employs creative linguistic variations.}
\label{tab:prompt_interaction}
\end{table}

\subsection{Post-Processing of Generated Tweets}
\label{sec:processing_tweets}

Following the generation of synthetic tweets by our language models, we merge them with human-written tweets and apply several post-processing steps to ensure the quality and consistency of the combined dataset. The primary goal of these steps is to remove low-quality or non-informative tweets while maintaining a consistent format.

We eliminated rows with empty text fields in the human-written and machine-generated tweets. After this, we normalized and cleaned the remaining text. Our normalization process includes expanding contractions, fixing special characters (e.g., Demojiizing emojis to their textual form), and removing unwanted encoding replacements. Tweets that are left empty after normalization are also removed. 

Next, we clean the tweets by removing those that do not primarily consist of text (e.g., tweets containing less than 50\% alphanumeric content). Additionally, tweets containing unintended phrases leaked from the model's prompt are identified and removed. These unintended phrases typically come from language model behaviors, such as chatbot responses. For example, common phrases include \say{you are an AI assistant,} \say{is there anything else I can help you with,} and \say{I cannot create explicit content.}

Randomly sampling 50 samples from each dataset, we found instances of AI-related hashtags that the models generated, often appearing in otherwise coherent tweets. Examples of such AI-generated hashtags include \#aichatbot, \#generativetweeting, \#aiwritestweets, and \#aibotassistant. Tweets containing less than three of these hashtags only have the hashtags removed; if there are more than three occurrences, we return an empty string, subsequently removing that sample. We also remove tweets that are too short to be meaningful (fewer than 10 characters) and remove any duplicate tweets that are identified to ensure data uniqueness. 

\begin{table}[h]
\centering
\scriptsize
\setlength{\tabcolsep}{4pt}
\begin{tabular}{@{}lcc@{}}
\toprule
\textbf{Model Family} & \textbf{Model} & \textbf{Rejection Rate (\%)} \\ \midrule
\multirow{3}{*}{Llama3} & LL3 & 4.94 \\
                       & LL3-Dolphin & 1.82 \\
                       & LL3-Hermes & 0.87 \\ \midrule
\multirow{3}{*}{Mistral} & Mistral & 2.30 \\
                         & Mistral-Dolphin & 1.96 \\
                         & Mistral-Hermes & 2.51 \\ \midrule
\multirow{2}{*}{Qwen} & Qwen2 & 12.31 \\
                      & Qwen2-Dolphin & 2.02 \\ \midrule
GPT4o & GPT4o & 0.14 \\ \midrule
Human & Human & 3.35 \\
\bottomrule
\end{tabular}
\caption{Rejection rates for human and machine-generated tweets by model family. Uncensored models generally have lower rejection rates than censored counterparts.}
\label{tab:rejection_rate_compare}
\end{table}

After completing these steps, the rejection rate is calculated as the percentage of the original dataset removed during post-processing. This is computed by comparing the total number of rows removed to the original dataset length:
\[
\frac{\text{Original Length} - \text{Processed Length}}{\text{Original Length}} \times 100
\]

We apply this method to both human and machine-generated tweets. Table \ref{tab:rejection_rate_compare} presents the final rejection rates.

\subsubsection{Human Vs. Machine-Generated Dataset Creation}
To evaluate the effectiveness of our fine-tuned language models in generating synthetic tweets, we construct datasets for each LLM variation using a $90/10/10$ train-validation-test split. Each dataset contains human-written and machine-generated tweets, which are paired to maintain consistency. This pairing ensures human-written and corresponding machine-generated tweets are kept within the same split (train, validation, or test) to prevent cross-contamination. For example, if a human-written tweet is in the training set, its machine-generated counterpart cannot be in the validation or test sets, and vice versa. This strict separation ensures that the models are not indirectly exposed to human-written and machine-generated versions of the same context across different data splits, preserving the integrity of the evaluation process and providing a fair assessment of each model’s ability to generalize to unseen content.

\begin{table}[h]
\centering
\scriptsize
\setlength{\tabcolsep}{4pt}
\begin{tabular}{lccc}
\toprule
\textbf{Split} & \textbf{Human (0)} & \textbf{Generated (1)} & \textbf{Total} \\ \midrule
Training & 3544 & 3544 & 7088 \\ 
Validation & 443 & 443 & 886 \\ 
Test & 443 & 443 & 886 \\ \midrule
Total & 4430 & 4430 & 8860 \\ 
\bottomrule
\end{tabular}
\caption{Distribution of train, validation, and test splits and corresponding label counts for each of the nine datasets.}
\label{tab:data-dist}
\end{table}

After constructing the datasets, we down-sample them to match the size of the smallest dataset, as some machine-generated tweets were excluded during post-processing at a higher rate due to quality issues (see section \ref{sec:processing_tweets}). Down-sampling ensures fairness in evaluation by eliminating potential biases from comparing datasets of unequal sizes, which might skew results in favor of larger datasets. This process ensures that performance differences reflect the content of the data rather than the dataset size. Throughout this process, the human-written tweets remain consistent across all datasets. Table \ref{tab:data-dist} provides details of the train, validation, test splits, and label distributions.

\subsection{Machine-Generated Text Detectors} We use the BERTweet-base model \cite{Nguyen2020BERTweetAP} and DeBERTaV3 base \cite{he2023debertav} as our initial baseline models, which generally show robust performance on Twitter-oriented downstream tasks. The models are fine-tuned on each of our nine datasets using the hyperparameters specified in Table \ref{tab:bertweet_hyperparams} in Appendix \ref{sec:appendixA}.

To further improve the performance and robustness of our machine-generated text detection system, we implement a soft voting ensemble of five BERTweet models and five DeBERTaV3 models. Each model in the ensemble is trained independently, only varying by initialization relying on random, using the same hyperparameters and dataset splits as the baseline model. During inference, the ensemble predicts the class probabilities for each input text, and the final prediction is determined by averaging the probabilities across all models and selecting the class with the highest average probability. 

Previous work \cite{fabien-etal-2020-bertaa} has shown that adding linguistic features concatenated with the \textit{CLS} token from bert-based models improves detection capabilities in authorship attribution. To test this on our nine datasets, we implement \cite{Kumarage2023StylometricDO} approach, utilizing a reduction network to gradually reduce the dimensionality of the \textit{CLS} $+$ stylometric features, followed by a final classification network to produce our probabilities of human or generated. The linguistic features integrated can be found in Table \ref{tab:stylometry-features} in appendix \ref{sec:appendixA}.

\begin{table*}[ht]
\centering
\resizebox{\textwidth}{!}{%
\begin{tabular}{lccc|ccc|ccc|ccc}
\toprule
 & \multicolumn{3}{c}{\textbf{Mean ± SD}} & \multicolumn{3}{c}{\textbf{Human vs Censored}} & \multicolumn{3}{c}{\textbf{Human vs Uncensored}} & \multicolumn{3}{c}{\textbf{Censored vs Uncensored}} \\
\cmidrule(lr){2-4}\cmidrule(lr){5-7}\cmidrule(lr){8-10}\cmidrule(lr){11-13}
\textbf{Metric} 
& \textbf{Human} & \textbf{Censored} & \textbf{Uncensored} 
& \textbf{p-val} & \textbf{d} & \textbf{CI (95\%)}  
& \textbf{p-val} & \textbf{d} & \textbf{CI (95\%)}  
& \textbf{p-val} & \textbf{d} & \textbf{CI (95\%)}  \\
\midrule
\multicolumn{13}{l}{\textbf{Lexical Richness}} \\[3pt]
Vocabulary Size 
& 15.65 ± 6.63 & 15.35 ± 6.86 & 15.62 ± 6.84
& 0.008** & 0.044 & [0.0817, 0.5213] 
& 0.863 & 0.003 & [-0.1924, 0.2375] 
& 9.53e-05*** & -0.040 & [-0.4143, -0.1437] \\[6pt]

MTTR 
& 0.948 ± 0.074 & 0.955 ± 0.085 & 0.951 ± 0.087 
& 2.75e-08*** & -0.092 & [-0.0099, -0.0049] 
& 0.006** & -0.043 & [-0.0060, -0.0010] 
& 1.87e-05*** & 0.044 & [0.0022, 0.0056] \\[6pt]

\midrule
\multicolumn{13}{l}{\textbf{Structural Patterns}} \\[3pt]
N-gram Diversity 
& 0.988 ± 0.053 & 0.987 ± 0.084 & 0.987 ± 0.089
& 0.355 & 0.014 & [-0.0010, 0.0030] 
& 0.292 & 0.015 & [-0.0009, 0.0031] 
& 0.884 & 0.001 & [-0.0016, 0.0018] \\[6pt]

N-gram Entropy 
& 3.49 ± 0.942 & 3.39 ± 1.01 & 3.44 ± 0.995
& 3.11e-09*** & 0.100 & [0.0664, 0.1294] 
& 0.001** & 0.055 & [0.0227, 0.0841] 
& 2.19e-05*** & -0.044 & [-0.0643, -0.0247] \\[6pt]

ISS 
& 0.860 ± 0.031 & 0.852 ± 0.030 & 0.859 ± 0.032
& 2.08e-53*** & 0.265 & [0.0072, 0.0093] 
& 0.002** & 0.051 & [0.0006, 0.0026] 
& 7.77e-96*** & -0.211 & [-0.0073, -0.0060] \\[6pt]

\midrule
\multicolumn{13}{l}{\textbf{Semantics}} \\[3pt]
BERTScore 
& — & 0.478 ± 0.061 & 0.476 ± 0.070
& — & — & — 
& — & — & — 
& 0.001** & 0.033 & [0.0009, 0.0035] \\
\bottomrule
\end{tabular}
}
\caption{Comparisons of Text Quality Metrics Across Conditions with Confidence Intervals. Mean ± SD values are reported for Human, Censored, and Uncensored conditions. Each pairwise comparison shows the p-value, Cohen’s d (effect size), and 95\% confidence intervals (CI) for the raw mean difference in the original units of the metric. Significance levels after adjustments: *** p<0.001, ** p<0.01, * p<0.05. BERTScore is not applicable to the Human condition as it already uses the Human and machine-generated pairs as candidate and reference sentences. ISS = Intra-sample Similarity; MTTR = Moving Average Type-Token Ratio. Censored and Uncensored indicate models with and without content filtering, respectively.}
\label{tab:combined_quality_analysis}
\end{table*}

\section{Results}
\label{sec:results}

We now present a detailed evaluation of nine large language model (LLM) variants, examining how different moderation strategies shape their linguistic properties and detectability. Our findings reveal that reducing moderation can improve a model’s linguistic fidelity to human text but come at the cost of increased toxicity and detection challenges. In the following subsections, we interpret the relationships of semantics, lexical richness, structure, toxicity, and detector performance to offer a nuanced perspective on the trade-offs introduced by censorship and its removal. 

Hyperparameter choices and LLM model links are in appendix \ref{sec:appendixA}. Our statistical significance testing guidelines are outlined in appendix \ref{sec:appendixB}. Appendix \ref{sec:appendixC} and \ref{sec:appendixD} comprehensively describe the metrics and complete individual results, respectively.

\subsection{Semantics, Lexical Richness, and Structural Patterns}
\label{sec:sem_lex_struc}

To understand how content moderation influences language generation quality, we examine three interrelated dimensions: lexical richness, structural patterns, and semantic fidelity (table \ref{tab:combined_quality_analysis}). These aspects help reveal how censorship constraints alter model outputs' diversity, internal coherence, and capacity to approximate human-like language use.

\subsubsection{Lexical Richness}

We first examine vocabulary size and the Moving Average Type-Token Ratio (MTTR) as measures of lexical richness. Although differences across conditions are statistically significant, they are generally accompanied by small effect sizes. In other words, moderation strategies lead to only subtle adjustments in word choice diversity rather than driving any major shifts.

Human and uncensored outputs demonstrate nearly identical vocabulary sizes, both slightly exceeding those of censored models. This outcome shows that relaxing restrictions does not substantially boost lexical richness; instead, it allows the model’s range of words to align more closely with human baselines. Similarly, while censored models maintain a marginally higher MTTR, the difference is minimal. Overall, these findings point to a modest influence of censorship on lexical breadth, gently nudging the model’s vocabulary usage rather than severely constraining it.

\subsubsection{Structural Patterns}

Next, we analyze the structural aspects of generated text through bigram diversity, bigram entropy, and intra-sample similarity (ISS). Unlike lexical richness, structural metrics reveal clearer patterns. Although bigram diversity remains largely consistent across human, censored, and uncensored texts, bigram entropy and ISS highlight meaningful distinctions.

Human-produced text exhibits slightly higher bigram entropy and ISS than censored outputs, reflecting a naturally balanced and cohesive arrangement of word pairs. In contrast, censored models show marginally more repetitive patterns, a likely consequence of restrictive moderation that diminishes structural variety. Uncensored models more closely approximate the human-like balance of entropy and coherence, narrowing the gap between constraint-driven repetition and the flexible patterning found in human language.

\begin{figure}[htbp]
    \centering
    \includegraphics[width=\columnwidth]{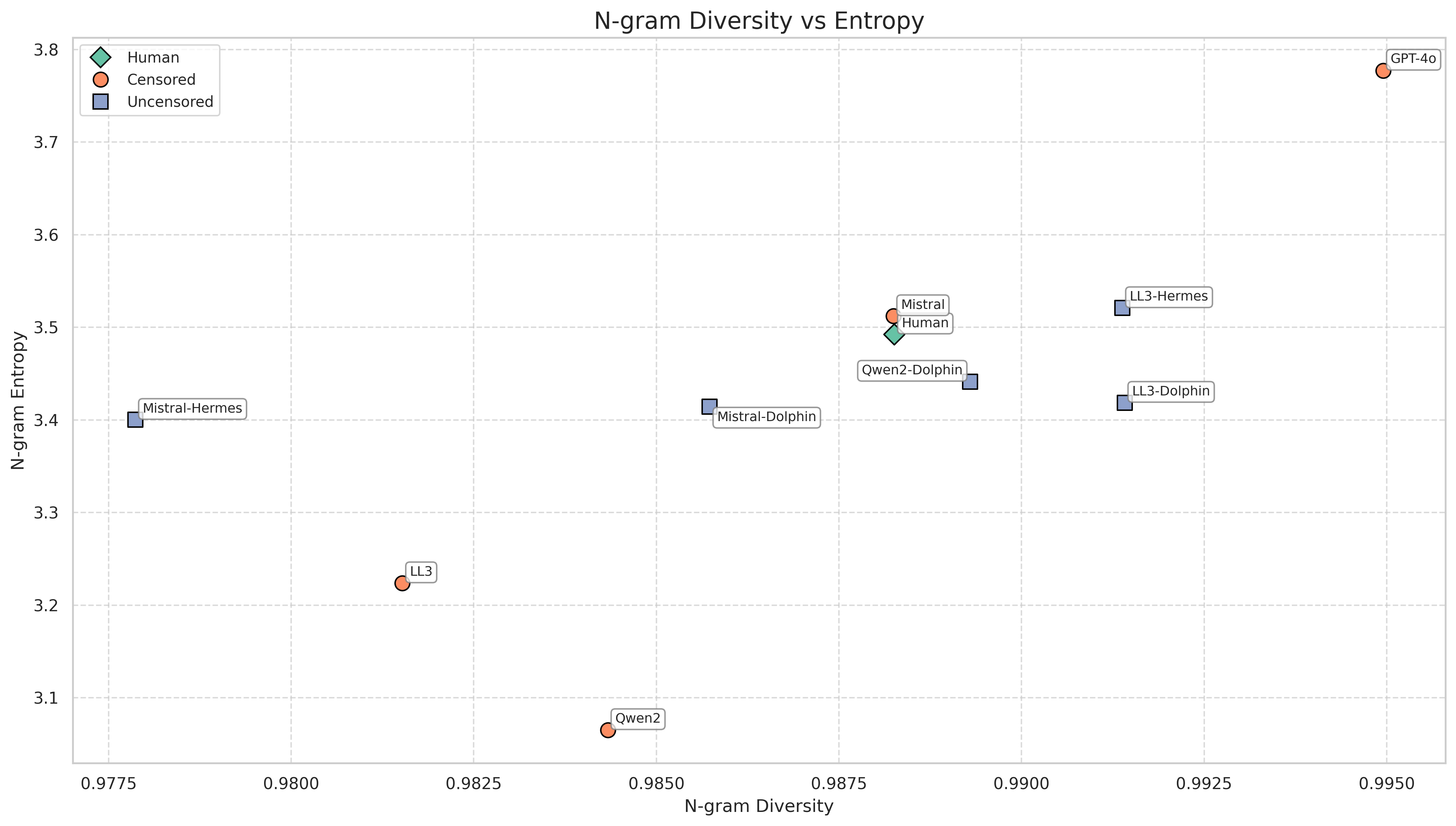}
    \caption{We plot each model and human text on n-gram diversity (x-axis) and entropy (y-axis). GPT-4o appears in the upper right, exceeding human complexity. Humans and some censored models (e.g., Mistral) cluster near the center, indicating balanced variety. Certain censored models (e.g., Qwen2) remain lower in diversity and entropy, while uncensored models trend upward, increasing complexity compared to censored variants but not always surpassing human levels.}
    \label{fig:ngram}
\end{figure}

Figure~\ref{fig:ngram} offers a visual perspective: While censored outputs cluster in regions of relatively lower entropy, uncensored models gravitate toward the human space, reflecting more intricate and less predictable lexical transitions. This result displays a central tension: constraints that reduce harmful or sensitive content may also dampen human language's natural complexity and variability.

\subsubsection{Semantics}

Turning to semantic similarity (BERTScore), both censored and uncensored texts remain moderately aligned with human references. Their subtle differences do not decisively favor one approach over the other. We find that moderation does not severely undermine the semantic integrity of generated text. Instead, censorship appears to create greater variability in semantic alignment—some censored outputs closely match human semantics, while others deviate more noticeably. Uncensored models, by contrast, maintain a more consistent semantic quality.

Considering lexical, structural, and semantic factors together reveals that easing strict moderation can help models regain human-like complexity and coherence. However, this comes at the cost of producing less filtered and variable content. The following section will explore the ethical implications of these shifts, especially in relation to toxicity and harmful content.

\subsection{Content Characteristics}
\label{sec:content_char}

\begin{figure}[h]
    \centering
    \includegraphics[width=\columnwidth]{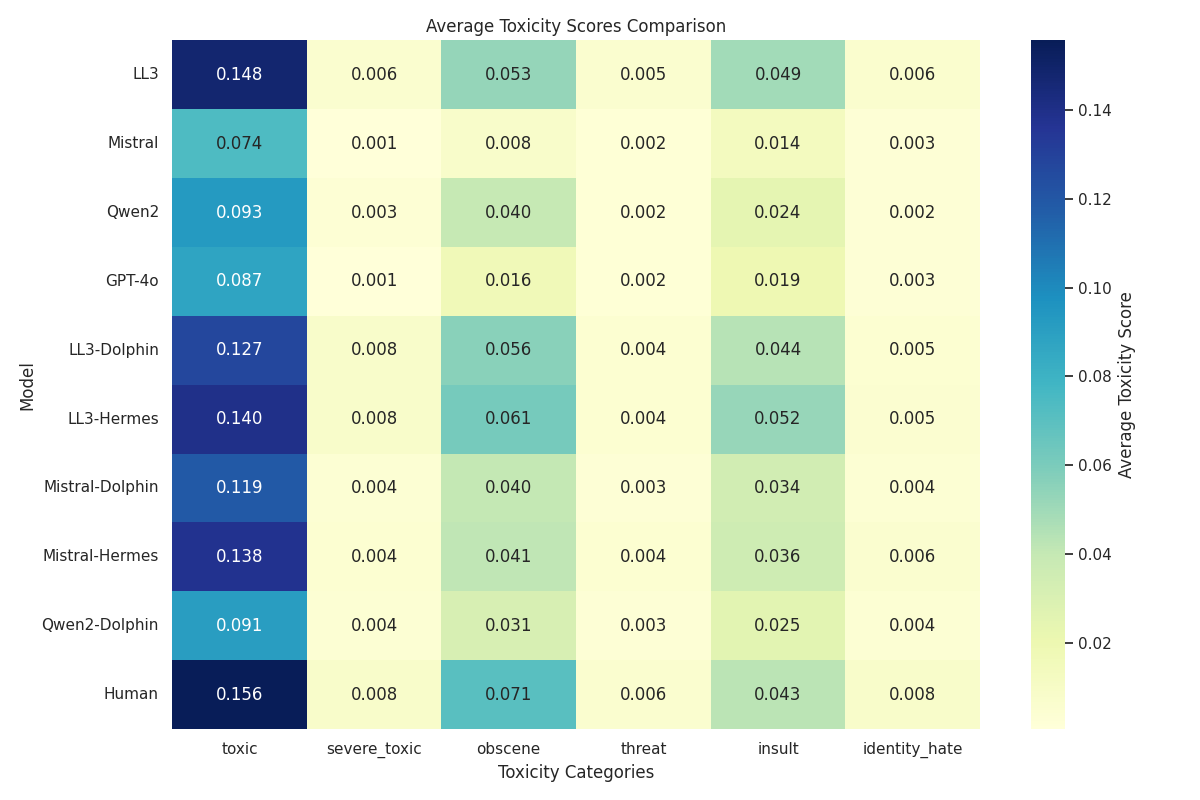}
    \caption{Comparison of fine-grained toxicity metrics. Each cell indicates the percentage of text samples exhibiting a given toxicity type. Human content is notably more toxic than most censored outputs, while some uncensored models approach human-level toxicity distributions.}
    \label{fig:toxicity}
\end{figure}

To explore the implications of censorship and its absence on the nature of generated text, we analyze fine-grained toxicity categories using the Toxic-BERT model \cite{Detoxify}. As shown in Figure~\ref{fig:toxicity}, human text frequently contains higher levels of toxic, insulting, or hateful content. Most censored models stay well below these toxicity baselines, showing that moderation measures effectively limit overtly harmful language.

However, certain censored models, such as LL3, break this pattern, showing that not all censored configurations consistently suppress toxicity. At the same time, many uncensored models exhibit higher toxicity levels, sometimes nearing those of human samples. Yet, some uncensored variants (e.g., Qwen2-Dolphin and LL3’s uncensored versions) remain comparatively less toxic. This outcome may reflect inherent stabilizing factors in their training or architecture, even without explicit filtering.

Overall, these toxicity results complement the earlier semantic and structural findings. While easing moderation can enhance a model’s complexity and resemblance to human-generated text, it also increases the potential for harmful content. Balancing these competing interests is essential for designing solutions that preserve linguistic richness while mitigating social and ethical risks associated with unrestricted text generation.

\begin{figure*}[h]
    \centering
    \includegraphics[width=\textwidth]{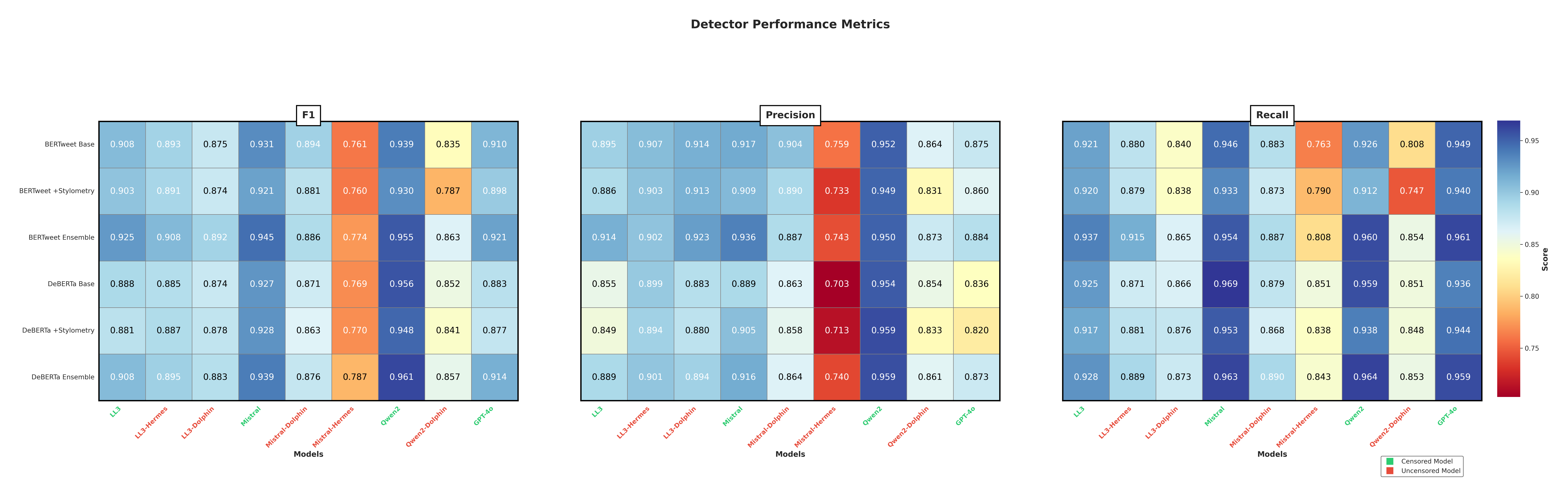} 
    \caption{Mean Precision, Recall, and F1 performance of our transformer-based detectors (BERTweet and DeBERTa variants, with and without stylometric features, and ensemble methods) tested across different censored and uncensored model outputs. Each heatmap cell represents the average metric score over five independent runs.}
    \label{fig:performance_comparison} 
\end{figure*}

\subsection{Detector Performance}
\label{sec:detector_perf}

Figure~\ref{fig:performance_comparison} presents our detector results across nine LLM variants. Three key insights emerge.

First, the Soft Ensemble detector (a combination of five BERTweet or DeBERTa models) consistently outperforms single-model approaches, achieving the highest F1 scores, precision, and recall. This increased stability of diverse initialization collectively offers a more robust decision boundary, mitigating individual model biases.

Second, attempts to augment BERTweet with stylometric features fail to deliver tangible gains. Adding stylometry slightly impairs performance compared to the plain BERTweet baseline. We find that stylometry captures subtle lexical and syntactic cues that do not provide incremental value once a strong neural baseline is established. As the generated text becomes more human-like, simple stylistic heuristics lose their discriminative power.

Third, and most critically, uncensored models significantly undermine detector effectiveness, especially on complex open-source variants like Mistral-Hermes and Qwen2-Dolphin. While censored models are still distinguishable from human text, removing moderation causes the generated content to mimic human linguistic patterns more closely. This is reflected in both precision and recall metrics:

\begin{itemize}
    \item \textit{Precision:} For censored models, precision remains relatively high. When detectors label a tweet as machine-generated, they tend to be correct. This reliability likely stems from the restricted stylistic and lexical patterns imposed by censorship, which make these outputs easier to distinguish from authentic human text. On the other hand, when moderation is lifted, precision declines. Freed from artificial constraints, uncensored models produce more nuanced and human-like language, causing detectors to misidentify an increasing amount of machine-generated tweets as genuine human content.
    
    \item \textit{Recall:} Uncensored outputs deal a more severe blow to recall. Since these models produce text nearly indistinguishable from human writing, detectors fail to identify a larger fraction of them as generated. The reduction in recall is pronounced for Mistral-Hermes and Qwen2-Dolphin, signifying that a substantial portion of their machine-generated tweets goes undetected.

\end{itemize}

It is important to note that all human samples remain identical across datasets, ensuring that performance declines are solely attributable to changes in the machine-generated subsets. Thus, this trend cannot be explained by a shift in human text; rather, it directly reflects the increasing human-likeness of uncensored model outputs.

\section{Conclusion}
\label{sec:conclusion}

This study provides a nuanced view of the trade-offs introduced by safety moderation in open-source large language models, particularly when generating domain-specific text. Our findings show that while uncensored models achieve greater lexical richness, structural complexity, and a closer semantic alignment to human language, these gains come with significant risks. Primarily, they elevate toxicity closer to human baselines, complicate detection efforts, and erode the safeguards intended to protect online discourse. In contrast, censored models maintain lower toxicity levels but produce less diverse and more predictable outputs, showcasing the delicate balance between mitigating harm and preserving linguistic richness.

Our results highlight the urgent need for more refined moderation strategies and sophisticated detection frameworks. Future research should seek moderation techniques that preserve a model’s creative potential without substantially increasing its capacity to produce harmful or deceptive content. Moreover, the diminishing effectiveness of current detectors against human-like, uncensored text emphasizes the importance of developing robust detection methods that can adapt to rapidly advancing generative capabilities.

\section{Limitations}
\label{sec:limitations}
Our study concentrates on Twitter data, and the resulting insights may not translate seamlessly to other social media platforms or domains. Twitter’s linguistic conventions—short utterances, hashtags, and dynamic engagement patterns—differ substantially from platforms like Reddit or Facebook, as well as from more formal text sources such as news articles and scientific publications. Future work should validate and refine our methods across diverse domains, languages, and evolving online communities.

We also rely on the TweetEval dataset for fine-tuning and evaluation. Although widely used, TweetEval may not fully capture real-world Twitter discourse's vast topical, linguistic, and demographic diversity. Social media language evolves rapidly, and models fine-tuned on static datasets risk becoming outdated. Periodic retraining on fresh samples and exploring additional datasets would help maintain detector robustness and broaden applicability.

Finally, our focus on smaller LLM parameter sizes (e.g., 7B and 8B) leaves open questions about the scalability and domain-transfer capabilities of substantially larger or smaller models. Future investigations could examine whether larger models produce more human-like content that evades detection or whether lightweight models trained on niche datasets could still maintain sufficient distinguishability. Exploring varying model sizes and training regimens will clarify the exchange between model capacity, content fidelity, and detectability.

\section{Ethical Considerations}
\label{sec:ethical}
Our work displays the dual-use nature of advanced language models. While their generative prowess can improve user experiences, it can also facilitate large-scale distribution of harmful or deceptive content. We acknowledge the risk that improved understanding of \textit{uncensored} LLMs might enable malicious actors to refine their tactics. To mitigate this, our focus is on advancing detection mechanisms, raising community awareness, and encouraging preemptive safeguards rather than enabling misuse.

The potential for biases and unfair outcomes also cannot be overlooked. Although we investigate models primarily on English Twitter data, real-world deployments must consider multilingual and multicultural settings. Ensuring that detectors do not disproportionately mislabel or penalize content from marginalized communities or sensitive contexts is of vast importance. Model developers, platform operators, and regulators must collaborate to define appropriate transparency standards, audit processes, and accountability frameworks.

Ultimately, the evolution of LLM detection systems and moderation policies must align with human rights, privacy, and equity principles. Balancing open scientific inquiry with responsible disclosure, our goal is to guide the ethical integration of these technologies into digital ecosystems, ensuring that innovations in language generation and detection serve the public good rather than undermine it.

\section*{Acknowledgments}

Research partly supported by NSF grants 2210198 and 2244279 and ARO grants W911NF-20-1-0254 and W911NF-23-1-0191. Verma is the founder of Everest Cyber Security and Analytics, Inc.

\bibliography{coling_latex}

\appendix

\section{Appendix A}
\label{sec:appendixA}

\subsection{Hyperparameters for Detectors and Stylometric Features}

\begin{table}[H]
\centering
\scriptsize
\setlength{\tabcolsep}{3pt}
\begin{tabular}{ll}
\toprule
\textbf{Hyperparameter} & \textbf{Value} \\
\midrule
Model & BERTweet \\
Epochs & 20 \\
Batch Size & 32 \\
Learning Rate & $2 \times 10^{-5}$ \\
Early Stopping & Validation Loss \\
Patience & 5 \\
Gradient Clipping & 1.0 \\
Warmup Proportion & 0.1 \\
$N$ Estimators (Ensemble) & 5 \\
\bottomrule
\end{tabular}
\caption{Training hyperparameters used for fine-tuning the BERTweet and DeBERTaV3 models, stylometric variations, and the ensemble detectors. These parameters were chosen based on best practices reported in prior work to balance computational efficiency and model stability.}
\label{tab:bertweet_hyperparams}
\end{table}

\begin{table}[H]
\centering
\scriptsize
\setlength{\tabcolsep}{3pt}
\begin{tabular}{ll}
\toprule
\textbf{Category} & \textbf{Features} \\ 
\midrule
Phraseology & Word \& sentence counts, paragraph counts \\
 & Mean word length \\
 & Mean/SD of words per sentence/paragraph \\
 & Mean/SD of sentences per paragraph \\
Punctuation & Total punctuation count \\
 & Counts of !, ?, :, ;, @, \# \\
Linguistic Diversity & Moving Avg. Type-Token Ratio (MTTR) \\
 & Readability (Flesch Reading Ease) \\
\bottomrule
\end{tabular}
\caption{Stylometric features integrated with BERT-based detectors to capture linguistic nuances beyond surface forms. These features include measures of phraseology, punctuation usage, and text complexity.}
\label{tab:stylometry-features}
\end{table}

\subsection{QLoRA Fine-Tuning Configuration}
\label{sec:qlora_finetune}

\begin{table}[H]
\centering
\scriptsize
\setlength{\tabcolsep}{3pt}
\begin{tabular}{l c}
\toprule
\textbf{Trainer Parameter} & \textbf{Value} \\
\midrule
Per Device Train Batch Size & 8 \\
Gradient Accumulation Steps & 4 \\
Optimizer & paged adamw 8bit \\
Learning Rate & 2e-4 \\
Weight Decay & 0.001 \\
Max Grad Norm & 1.0 \\
Max Steps & 2,856 \\
Warmup Ratio & 0.05 \\
LR Scheduler & cosine \\
FP16 & True \\
\bottomrule
\end{tabular}
\caption{Trainer configuration parameters for QLoRA-based fine-tuning of the language models. These settings were determined through pilot experiments and literature-recommended defaults, aiming to maintain training efficiency, numerical stability, and smooth convergence when adapting the base instruct models to the Twitter domain.}
\label{tab:trainer_params}
\end{table}

\begin{table}[h]
\centering
\scriptsize
\setlength{\tabcolsep}{3pt}
\begin{tabular}{l c}
\toprule
\textbf{LoRA Parameter} & \textbf{Value} \\
\midrule
R & 16 \\
LoRA Alpha & 32 \\
Target Modules & q,k,v,o,gate,up,down proj \\
Bias & none \\
LoRA Dropout & 0.05 \\
Task Type & CAUSAL LM \\
\bottomrule
\end{tabular}
\caption{LoRA configuration used for parameter-efficient fine-tuning. Adjusting the rank (R) and dropout parameters allows for retaining model performance while significantly reducing computational overhead.}
\label{tab:lora_params}
\end{table}

\subsection{Censored and Uncensored Model Links}
\label{sec:model_links}

\begin{table}[H]
    \centering
    \begin{adjustbox}{max width=\columnwidth}
    \begin{tabular}{|l|l|}
    \hline
    \textbf{Model} & \textbf{Footnote/Link} \\ \hline
    Meta-Llama-3-8B-Instruct & \url{https://huggingface.co/meta-llama/Meta-Llama-3-8B-Instruct} \\ \hline
    Hermes 2 Pro-Llama-3-8B & \url{https://huggingface.co/NousResearch/Hermes-2-Pro-Llama-3-8B} \\ \hline
    Dolphin-2.9-Llama-3-8B & \url{https://huggingface.co/cognitivecomputations/dolphin-2.9-llama3-8b} \\ \hline
    Mistral-7B-Instruct-v0.2 & \url{https://huggingface.co/Mistralai/Mistral-7B-Instruct-v0.2} \\ \hline
    Dolphin-2.8-Mistral-7b-v02 & \url{https://huggingface.co/cognitivecomputations/dolphin-2.8-mistral-7b-v02} \\ \hline
    OpenHermes-2.5-Mistral-7B & \url{https://huggingface.co/teknium/OpenHermes-2.5-Mistral-7B} \\ \hline
    Qwen2-7B-Instruct & \url{https://huggingface.co/Qwen/Qwen2-7B-Instruct} \\ \hline
    Dolphin-2.9.2-Qwen2-7B & \url{https://huggingface.co/cognitivecomputations/Dolphin-2.9.2-qwen2-7b} \\ \hline
    \end{tabular}
    \end{adjustbox}
    \caption{Links to the open-source models used in this study. These repositories provide model weights and documentation.}
    \label{tab:model_links}
\end{table}

\section{Appendix B}
\label{sec:appendixB}

\subsection{Statistical Testing} \label{sec:stats_testing}

In our evaluation, we categorized our data into three groups: \textit{human}, \textit{censored}, and \textit{uncensored}. We conducted pairwise comparisons to assess differences in each metric among these groups (i.e., human vs. censored, human vs. uncensored, and censored vs. uncensored).

With large sample sizes (approximately $4430$ human, $17720$ censored, and $22150$ uncensored instances), standard normality and variance homogeneity tests (e.g., Shapiro-Wilk, Levene’s test) become overly sensitive and can flag even negligible deviations as statistically significant \citep{lin2013}. Since the Central Limit Theorem ensures that the sampling distribution of the mean approximates normality for large samples, we did not explicitly test for distributional assumptions.

Our statistical testing protocol was as follows:

\begin{enumerate} 
    \item \textbf{Welch’s t-test:} We employed Welch’s t-test for each pairwise comparison to evaluate whether mean differences were statistically significant. Welch’s t-test is well-suited to large and potentially heterogeneous datasets, as it does not assume equal variances or equal sample sizes \citep{welch1947}.
    
    \item \textbf{Effect Size (Cohen’s d):} Beyond p-values, we computed Cohen’s d to gauge the practical importance of observed differences. While statistical significance can be influenced by large sample sizes, effect size metrics provide additional insight into the magnitude of differences \citep{cohen1988}.

    \item \textbf{Confidence Intervals:} We also derived 95\% confidence intervals for mean differences to quantify our estimates' precision and help visualize the range of plausible effect sizes.

    \item \textbf{Multiple Comparison Adjustment:} Given multiple pairwise comparisons, we applied the Benjamini-Hochberg procedure to control the False Discovery Rate (FDR). This step mitigates the risk of inflating Type I errors (false positives) due to multiple testing \citep{benjamini1995}.
\end{enumerate}

Our approach ensures that conclusions drawn from these comparisons are both statistically and practically meaningful, especially in the context of large-scale datasets.

\section{Appendix C}
\label{sec:appendixC}

\subsection{Metric Formulas}
\label{sec:metric_forumlas}

\subsubsection{BERTScore}
\label{sec:bertscore}
BERTScore~\cite{Zhang2020BERTScore:} evaluates the similarity between a candidate sentence \( C \) and a reference sentence \( R \) using contextual embeddings from a pre-trained language model (e.g., BERT).
We use the F1 metric using the official implementation found at \url{https://github.com/Tiiiger/bert_score}.

\subsubsection{Vocabulary Size}
\label{sec:vocab_size}
Vocabulary size measures the richness of the language in the corpus. A larger vocabulary suggests more diverse word usage, while a smaller vocabulary indicates more repetition or simpler language structures.

The vocabulary size is the number of unique words in the corpus:
\begin{equation}
\text{Vocabulary Size} = \left| \{\text{unique words}\} \right|
\end{equation}

\subsubsection{Moving Type-Token Ratio (MTTR)}
\label{sec:mttr}
The Moving Type-Token Ratio (MTTR) measures lexical variety by analyzing the diversity of words within a sliding window of fixed size across a text. Unlike the traditional TTR, which considers the entire text, MTTR provides a more granular view of language diversity by calculating the Type-Token Ratio (TTR) within each window and then averaging these values. Higher MTTR values indicate consistently diverse language use across different segments of the text, while lower values suggest repetitive language within the windows.

The MTTR is calculated as follows:
\begin{equation}
\text{MTTR} = \frac{1}{N} \sum_{i=1}^{N} \frac{\left| \{\text{unique words in window } i\} \right|}{w}
\end{equation}
where:
\begin{itemize}
    \item \( w \) is the fixed window size (e.g., 10 words).
    \item \( N \) is the total number of windows, determined by the length of the text and the window size.
    \item \( \left| \{\text{unique words in window } i\} \right| \) represents the number of unique words within the \( i \)-th window.
\end{itemize}

\noindent
If the total number of words in the text is less than the window size \( w \), the MTTR is equivalent to the traditional TTR:
\begin{equation}
\text{MTTR} = \frac{\left| \{\text{unique words}\} \right|}{\text{Total Number of Words}}
\end{equation}

\noindent
This approach allows for a localized assessment of lexical diversity, providing insights into how word variety fluctuates throughout the text.

\subsubsection{N-gram Diversity}
\label{sec:ngram_divers}
N-gram diversity quantifies how varied the sequences of words (n-grams) are in the corpus. A higher value suggests more unique word combinations, while a lower value indicates repeated word usage patterns.

N-gram diversity is calculated as:
\begin{equation}
\text{N-gram Diversity} = \frac{\left| \{\text{unique n-grams}\} \right|}{\text{Total Number of n-grams}}
\end{equation}

\paragraph{Note}
\begin{itemize}
    \item We evaluate bigrams in our work.
\end{itemize}

\subsubsection{N-gram Entropy}
\label{sec:ngram_ent}
N-gram entropy measures the unpredictability or randomness of word sequences in the text. Higher entropy reflects more diverse word combinations, while lower entropy suggests more predictable, repetitive patterns.

N-gram entropy is calculated using Shannon's entropy formula:
\begin{equation}
\text{N-gram Entropy} = -\sum_{i=1}^{N} p_i \log_2(p_i)
\end{equation}
where \( p_i \) is the probability of the \( i \)-th n-gram, defined as:
\begin{equation}
p_i = \frac{\text{Frequency of n-gram } i}{\text{Total Number of n-grams}}
\end{equation}

\paragraph{Note}
\begin{itemize}
    \item We evaluate bigrams in our work.
\end{itemize}

\subsubsection{Intra-sample Similarity}
\label{sec:iss}
Intra-sample similarity assesses how coherent or repetitive a sample is by comparing the similarity of word embeddings within each sentence. High similarity indicates sentences with closely related words, while lower similarity suggests more varied or unrelated content. This metric can highlight the degree of internal coherence within a sample.

It is calculated as follows:

\begin{equation}
\text{Intra-sample Similarity} = \frac{N}{\sum_{i=1}^{N} \left| S_i - \theta \right|}
\end{equation}

\noindent where:
\begin{itemize}
    \item $N$ is the total number of sentences in the sample.
    \item $S_i$ is the average cosine similarity between all pairs of words in the $i$-th sentence.
    \item $\theta$ is the minimum word similarity threshold (default value: 0.5).
\end{itemize}

The average cosine similarity for a sentence is computed as:

\begin{equation}
S_i = \frac{2}{M(M-1)} \sum_{j=1}^{M} \sum_{k=j+1}^{M} \cos(\mathbf{w}_j, \mathbf{w}_k)
\end{equation}

\noindent where:
\begin{itemize}
    \item $M$ is the number of words in the sentence.
    \item $\mathbf{w}_j$ and $\mathbf{w}_k$ are the word embeddings for the $j$-th and $k$-th words, respectively.
    \item $\cos(\mathbf{w}_j, \mathbf{w}_k)$ denotes the cosine similarity between the word embeddings.
\end{itemize}

\paragraph{Note}

\begin{itemize}
    \item The factor $\frac{2}{M(M-1)}$ in the average cosine similarity formula accounts for the number of unique word pairs in a sentence of length $M$.
    \item Word embeddings ($\mathbf{w}_j$, $\mathbf{w}_k$) are computed using BERTweet.
    \item Sentences containing only one word are excluded from similarity calculations, as cosine similarity requires at least two word vectors.
\end{itemize}

\section{Appendix D}
\label{sec:appendixD}
\subsection{Semantic Meaning, Lexical Richness, and Structural Patterns Individual Model Results}
\label{sec:app_individual_results}

\subsubsection{BERTScore}

\begin{figure}[H]
    \centering
    \includegraphics[width=\columnwidth]{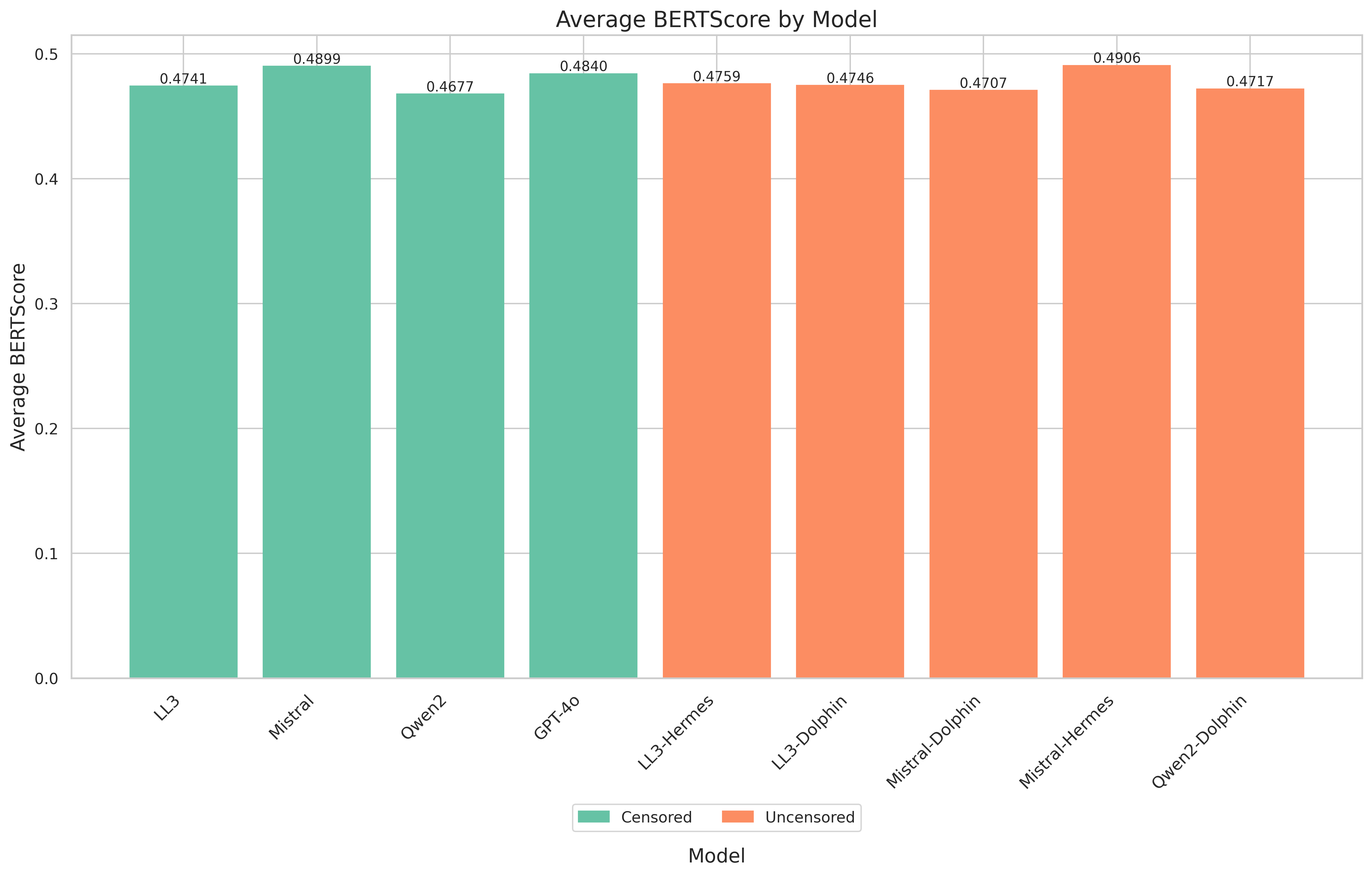}
    \caption{Average BERTScore across different censored and uncensored model variants. 
    Higher BERTScore values indicate that model outputs are semantically more aligned with the human reference texts. 
    The results show minimal differences in overall semantic fidelity among models, with both censored and uncensored versions achieving scores close to one another. 
    This suggests that while removing moderation constraints may slightly shift linguistic patterns, it does not substantially degrade semantic alignment with the original tweets.}
    \label{fig:bertsc_indv}
\end{figure}

\subsubsection{Vocabulary Size}

\begin{figure}[H]
    \centering
    \includegraphics[width=\columnwidth]{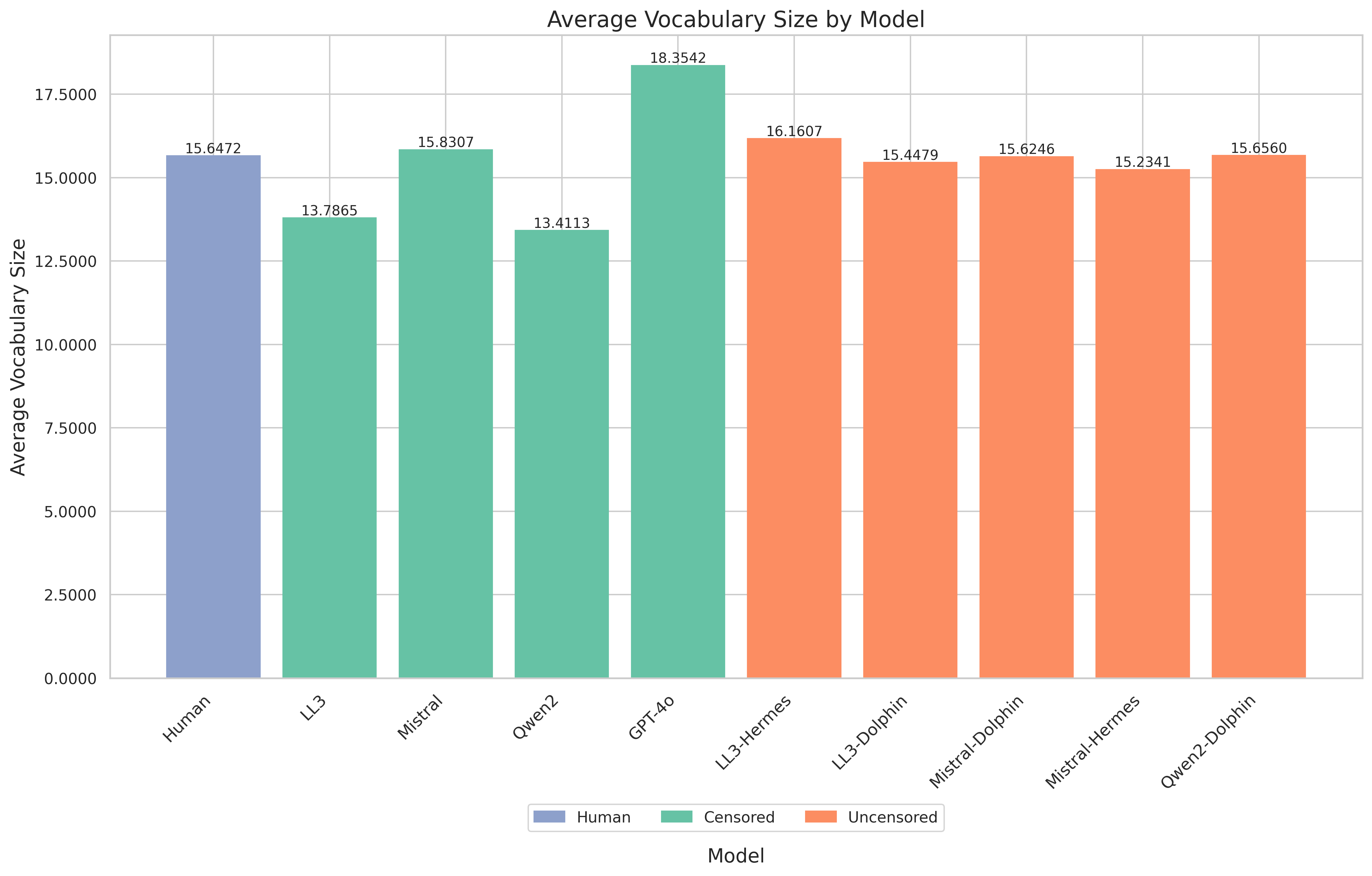}
    \caption{Average vocabulary size measured across human-written texts, censored models, and uncensored models. While some censored models (e.g., Qwen2) show a smaller lexicon, others like GPT-4o exceed the human baseline, indicating greater lexical diversity. Uncensored models generally fall closer to human-level vocabulary sizes, narrowing the gap in lexical variety.}
    \label{fig:voc_indv}
\end{figure}

\subsubsection{MTTR}

\begin{figure}[H]
    \centering
    \includegraphics[width=\columnwidth]{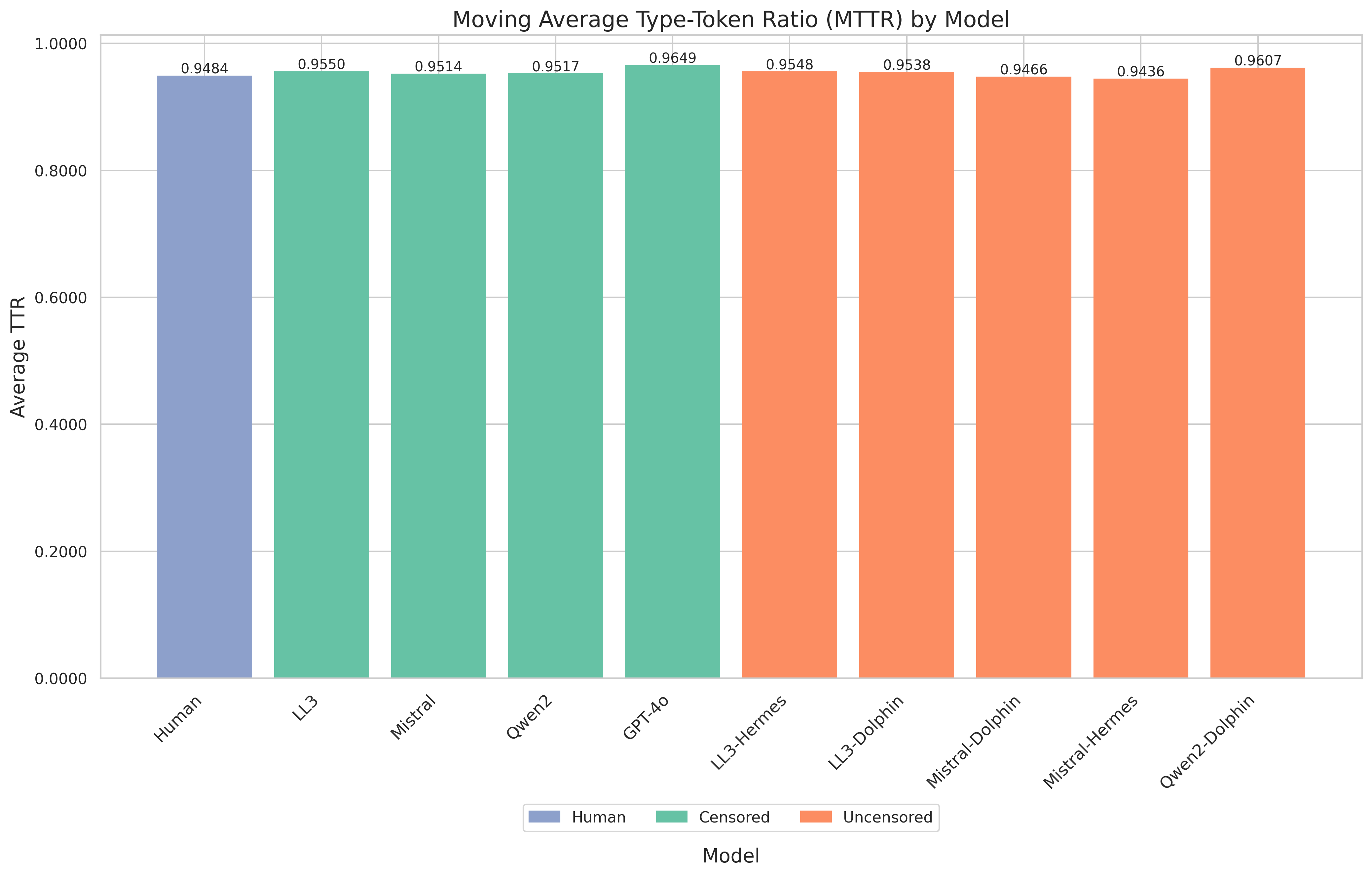}
    \caption{Comparison of the Moving Average Type-Token Ratio (MTTR) across human-written tweets, censored models, and uncensored models. MTTR values are all quite close, with both censored and uncensored outputs showing levels of lexical variability on par with human samples. Our results indicate that adjustments to moderation do not markedly diminish or enhance the consistency of word diversity when measured over small text windows, preserving a largely stable distribution of word usage patterns.}
    \label{fig:ttr_indv}
\end{figure}

\subsubsection{N-gram Diversity}

\begin{figure}[H]
    \centering
    \includegraphics[width=\columnwidth]{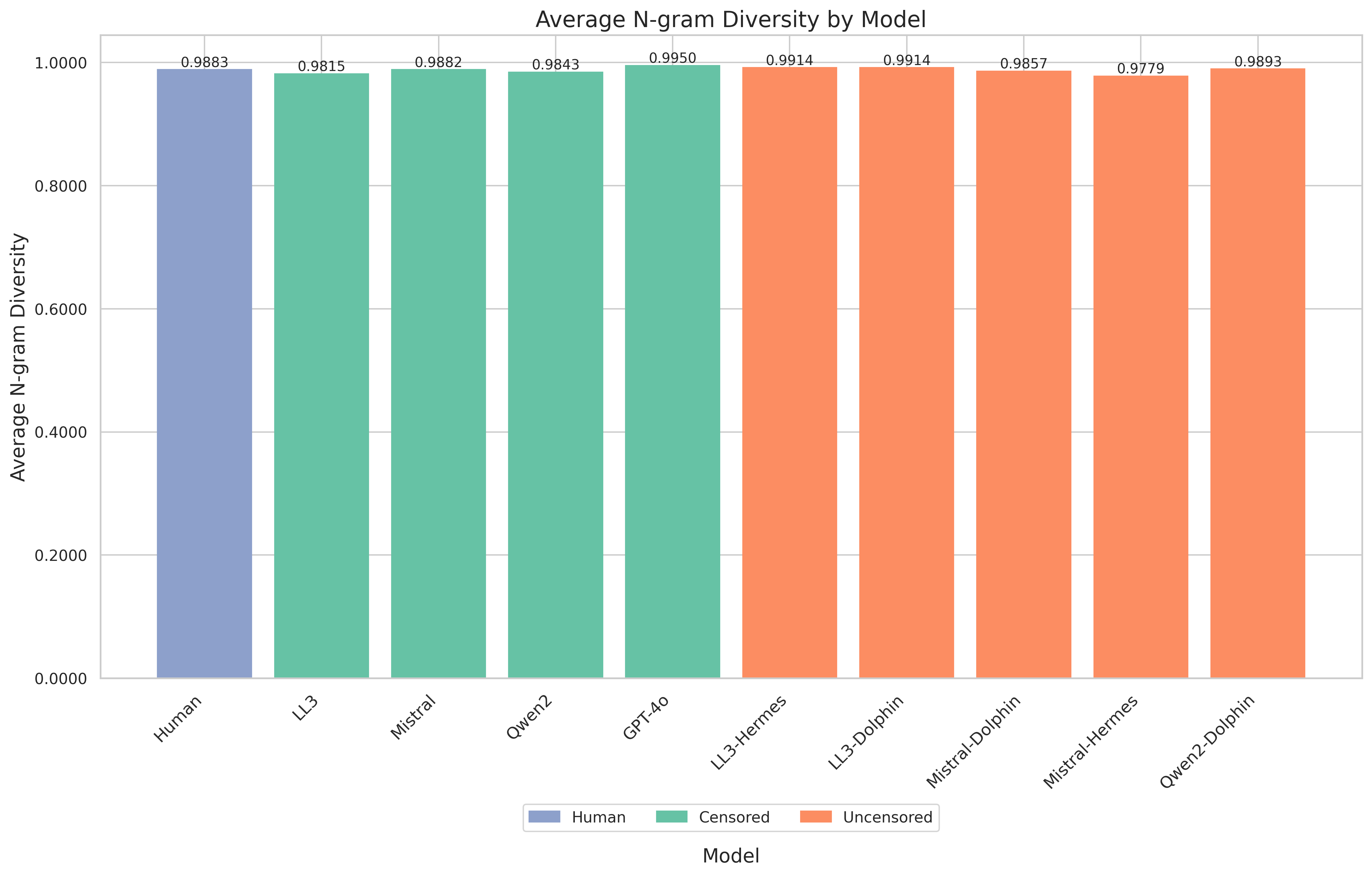}
    \caption{Average n-gram diversity across human-written text, censored models, and uncensored models. 
    All values cluster close to one, indicating a high level of word combination variety regardless of moderation status. While some uncensored models and GPT-4o slightly exceed the human baseline, and certain censored models fall just below it, these differences are marginal.}
    \label{fig:ngram_div_indv}
\end{figure}

\subsubsection{N-Gram Entropy}

\begin{figure}[H]
    \centering
    \includegraphics[width=\columnwidth]{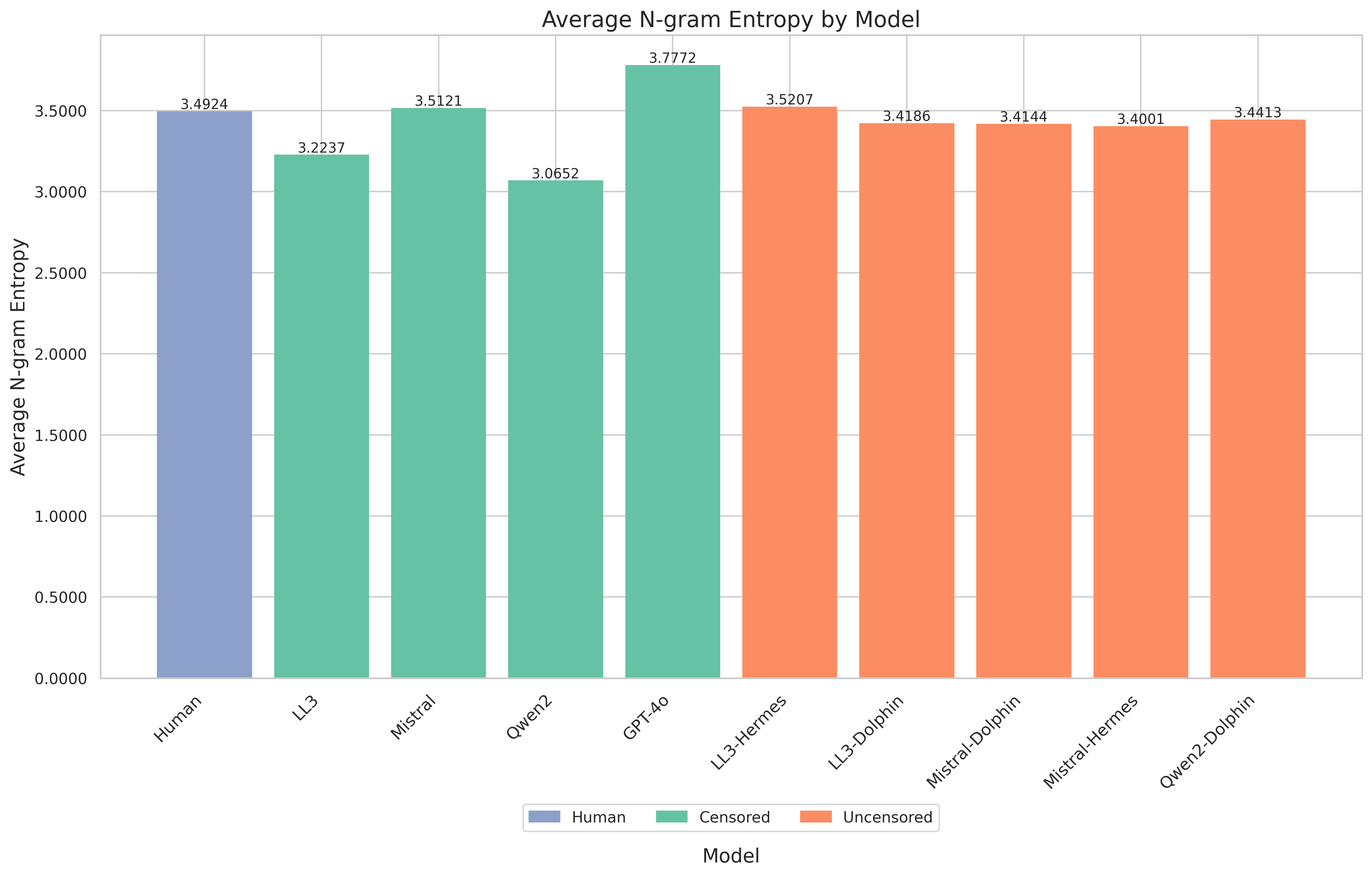}
    \caption{Average n-gram entropy across human-written text, censored models, and uncensored models. 
    N-gram entropy measures the unpredictability and complexity of word sequences: higher values indicate more varied and less predictable patterns. Most censored models exhibit lower entropy than human-generated text, reflecting more repetitive sequences. In contrast, several uncensored models rise to or exceed human-level entropy, indicating that removing moderation constraints fosters more linguistically diverse and unpredictable text generation.}
    \label{fig:ngram_entr_indv}
\end{figure}

\subsubsection{Intra-sample Similarity}

\begin{figure}[H]
    \centering
    \includegraphics[width=\columnwidth]{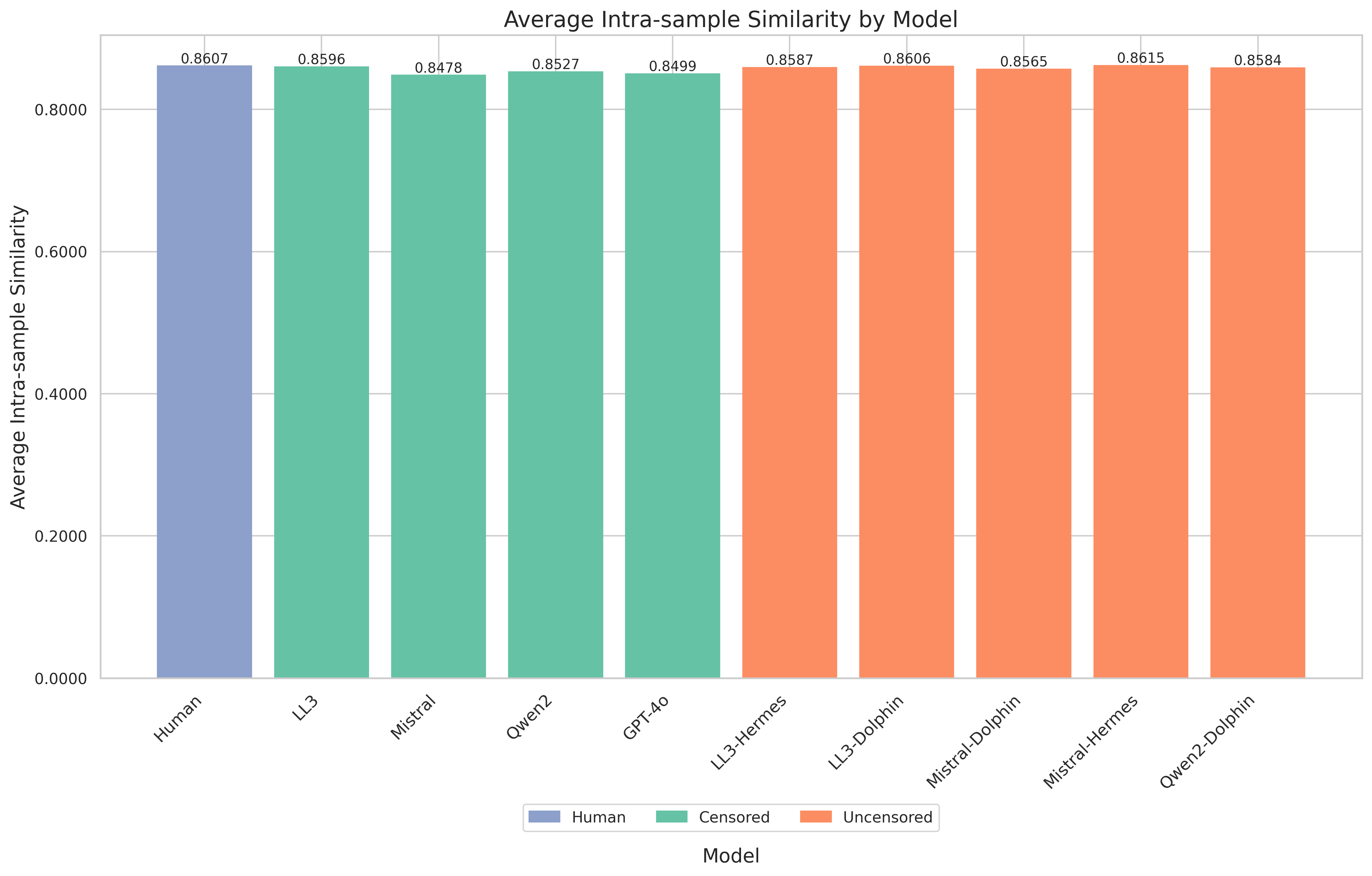}
    \caption{Average intra-sample similarity across human-written texts, censored models, and uncensored models. Intra-sample similarity quantifies how closely related words and sentences are within a single text sample. Although the values are relatively high across all conditions, human-written tweets display slightly higher similarity, suggesting a more cohesive and contextually interlinked structure.}
    \label{fig:iss_indv}
\end{figure}

\subsubsection{Detector Performance}

\begin{table*}[htbp]
\centering
\footnotesize
\resizebox{\textwidth}{!}{%
\begin{tabular}{llccccc}
\toprule
Dataset & Model & Precision & Recall & F1 Score & Accuracy & MCC \\
\midrule
\multirow{6}{*}{LL3}
& Bertweet & 0.895±0.011 & 0.921±0.008 & 0.908±0.005 & 0.907±0.005 & 0.814±0.010 \\
& DeBERTa & 0.855±0.015 & 0.925±0.020 & 0.888±0.004 & 0.884±0.004 & 0.771±0.008 \\
& Bertweet + Stylo & 0.886±0.015 & 0.920±0.008 & 0.903±0.006 & 0.901±0.007 & 0.802±0.013 \\
& DeBERTa + Stylo & 0.849±0.017 & 0.917±0.028 & 0.882±0.011 & 0.877±0.011 & 0.757±0.022 \\
& Bertweet Ensemble & \textbf{0.914±0.005} & \textbf{0.937±0.005} & \textbf{0.925±0.003} & \textbf{0.924±0.003} & \textbf{0.849±0.006} \\
& DeBERTa Ensemble & 0.889±0.021 & 0.928±0.031 & 0.908±0.015 & 0.906±0.015 & 0.813±0.030 \\
\midrule
\multirow{6}{*}{LL3-Hermes}
& Bertweet & \textbf{0.907±0.014} & 0.880±0.015 & 0.893±0.008 & 0.895±0.008 & 0.790±0.017 \\
& DeBERTa & 0.899±0.007 & 0.871±0.022 & 0.885±0.009 & 0.886±0.007 & 0.774±0.013 \\
& Bertweet + Stylo & 0.903±0.009 & 0.879±0.018 & 0.891±0.008 & 0.892±0.007 & 0.785±0.014 \\
& DeBERTa + Stylo & 0.894±0.012 & 0.881±0.024 & 0.887±0.016 & 0.888±0.014 & 0.776±0.029 \\
& Bertweet Ensemble & 0.902±0.006 & \textbf{0.915±0.010} & \textbf{0.909±0.005} & \textbf{0.908±0.005} & \textbf{0.816±0.010} \\
& DeBERTa Ensemble & 0.901±0.015 & 0.889±0.021 & 0.895±0.011 & 0.896±0.010 & 0.791±0.020 \\
\midrule
\multirow{6}{*}{LL3-Dolphin}
& Bertweet & 0.914±0.010 & 0.840±0.022 & 0.875±0.011 & 0.880±0.009 & 0.763±0.016 \\
& DeBERTa & 0.883±0.016 & 0.866±0.015 & 0.874±0.006 & 0.875±0.006 & 0.751±0.012 \\
& Bertweet + Stylo & 0.913±0.007 & 0.838±0.014 & 0.874±0.008 & 0.879±0.007 & 0.761±0.013 \\
& DeBERTa + Stylo & 0.880±0.005 & \textbf{0.876±0.012} & 0.878±0.006 & 0.878±0.005 & 0.757±0.011 \\
& Bertweet Ensemble & \textbf{0.924±0.024} & 0.865±0.032 & \textbf{0.892±0.009} & \textbf{0.896±0.007} & \textbf{0.795±0.012} \\
& DeBERTa Ensemble & 0.894±0.005 & 0.873±0.008 & 0.883±0.003 & 0.885±0.003 & 0.770±0.006 \\
\midrule
\multirow{6}{*}{Mistral}
& Bertweet & 0.917±0.014 & 0.946±0.006 & 0.931±0.005 & 0.930±0.006 & 0.861±0.011 \\
& DeBERTa & 0.889±0.013 & \textbf{0.969±0.009} & 0.927±0.007 & 0.924±0.008 & 0.852±0.015 \\
& Bertweet + Stylo & 0.909±0.011 & 0.933±0.015 & 0.921±0.005 & 0.920±0.005 & 0.840±0.009 \\
& DeBERTa + Stylo & 0.905±0.013 & 0.953±0.012 & 0.929±0.006 & 0.927±0.006 & 0.855±0.012 \\
& Bertweet Ensemble & \textbf{0.936±0.006} & 0.954±0.004 & \textbf{0.945±0.002} & \textbf{0.945±0.002} & \textbf{0.890±0.005} \\
& DeBERTa Ensemble & 0.916±0.009 & 0.963±0.009 & 0.939±0.002 & 0.937±0.003 & 0.875±0.005 \\
\midrule
\multirow{6}{*}{Mistral-Dolphin}
& Bertweet & \textbf{0.904±0.013} & 0.884±0.011 & \textbf{0.894±0.010} & \textbf{0.895±0.010} & \textbf{0.790±0.020} \\
& DeBERTa & 0.863±0.009 & 0.879±0.013 & 0.871±0.007 & 0.870±0.007 & 0.739±0.015 \\
& Bertweet + Stylo & 0.890±0.005 & 0.873±0.019 & 0.881±0.010 & 0.882±0.009 & 0.765±0.017 \\
& DeBERTa + Stylo & 0.858±0.018 & 0.868±0.010 & 0.863±0.005 & 0.862±0.007 & 0.724±0.014 \\
& Bertweet Ensemble & 0.887±0.019 & 0.887±0.015 & 0.887±0.003 & 0.886±0.005 & 0.773±0.010 \\
& DeBERTa Ensemble & 0.864±0.010 & \textbf{0.890±0.020} & 0.876±0.005 & 0.875±0.003 & 0.750±0.007 \\
\midrule
\multirow{6}{*}{Mistral-Hermes}
& Bertweet & \textbf{0.759±0.010} & 0.763±0.023 & 0.761±0.009 & 0.760±0.006 & 0.521±0.012 \\
& DeBERTa & 0.703±0.019 & \textbf{0.851±0.020} & 0.770±0.012 & 0.745±0.016 & 0.502±0.029 \\
& Bertweet + Stylo & 0.733±0.008 & 0.790±0.013 & 0.760±0.007 & 0.751±0.006 & 0.504±0.013 \\
& DeBERTa + Stylo & 0.713±0.027 & 0.838±0.018 & 0.770±0.014 & 0.750±0.021 & 0.502±0.037 \\
& Bertweet Ensemble & 0.743±0.020 & 0.808±0.023 & 0.774±0.003 & 0.764±0.008 & 0.531±0.013 \\
& DeBERTa Ensemble & 0.740±0.026 & 0.843±0.021 & \textbf{0.787±0.008} & \textbf{0.772±0.014} & \textbf{0.551±0.025} \\
\midrule
\multirow{6}{*}{Qwen2}
& Bertweet & 0.952±0.010 & 0.926±0.012 & 0.939±0.004 & 0.940±0.003 & 0.880±0.007 \\
& DeBERTa & 0.954±0.013 & 0.959±0.017 & 0.956±0.007 & 0.956±0.006 & 0.912±0.012 \\
& Bertweet + Stylo & 0.949±0.007 & 0.912±0.017 & 0.930±0.008 & 0.932±0.007 & 0.864±0.014 \\
& DeBERTa + Stylo & \textbf{0.959±0.010} & 0.938±0.018 & 0.948±0.006 & 0.949±0.005 & 0.898±0.010 \\
& Bertweet Ensemble & 0.950±0.006 & 0.960±0.005 & 0.955±0.003 & 0.954±0.003 & 0.909±0.006 \\
& DeBERTa Ensemble & \textbf{0.959±0.015} & \textbf{0.964±0.005} & \textbf{0.961±0.006} & \textbf{0.961±0.006} & \textbf{0.922±0.012} \\
\midrule
\multirow{6}{*}{Qwen2-Dolphin}
& Bertweet & 0.864±0.012 & 0.808±0.027 & 0.835±0.011 & 0.840±0.008 & 0.682±0.015 \\
& DeBERTa & 0.854±0.015 & 0.851±0.023 & 0.852±0.008 & 0.852±0.007 & 0.705±0.014 \\
& Bertweet + Stylo & 0.831±0.068 & 0.747±0.050 & 0.787±0.058 & 0.797±0.057 & 0.598±0.115 \\
& DeBERTa + Stylo & 0.833±0.011 & 0.848±0.009 & 0.841±0.009 & 0.839±0.010 & 0.678±0.019 \\
& Bertweet Ensemble & \textbf{0.873±0.011} & \textbf{0.854±0.016} & \textbf{0.863±0.003} & \textbf{0.865±0.002} & \textbf{0.730±0.003} \\
& DeBERTa Ensemble & 0.861±0.019 & 0.853±0.013 & 0.857±0.007 & 0.857±0.008 & 0.715±0.016 \\
\midrule
\multirow{6}{*}{GPT4o}
& Bertweet & 0.875±0.008 & 0.949±0.006 & 0.910±0.003 & 0.907±0.003 & 0.816±0.006 \\
& DeBERTa & 0.836±0.028 & 0.936±0.032 & 0.883±0.006 & 0.876±0.008 & 0.759±0.013 \\
& Bertweet + Stylo & 0.860±0.015 & 0.940±0.012 & 0.898±0.007 & 0.893±0.008 & 0.790±0.015 \\
& DeBERTa + Stylo & 0.820±0.016 & 0.944±0.015 & 0.877±0.004 & 0.868±0.006 & 0.745±0.008 \\
& Bertweet Ensemble & \textbf{0.884±0.013} & \textbf{0.961±0.006} & \textbf{0.921±0.007} & \textbf{0.917±0.008} & \textbf{0.838±0.014} \\
& DeBERTa Ensemble & 0.873±0.004 & 0.959±0.007 & 0.914±0.002 & 0.910±0.002 & 0.824±0.005 \\
\bottomrule
\end{tabular}
} 
\caption{Comparison of BERTweet, DeBERTa, and their ensemble and stylometry-augmented variants across multiple datasets. Metrics include Precision, Recall, F1 Score, Accuracy, and Matthews Correlation Coefficient (MCC), averaged over five random seeds. Bold values highlight the best performance for each dataset-metric pair.}
\label{tab:full_performance_comparison}
\end{table*}

\end{document}